\documentclass{article} %
\usepackage{iclr2026_conference,times}

\usepackage{amsmath,amsfonts,bm}

\def\eqref#1{equation~\ref{#1}}

\def\1{\bm{1}}

\DeclareMathAlphabet{\mathsfit}{\encodingdefault}{\sfdefault}{m}{sl}
\SetMathAlphabet{\mathsfit}{bold}{\encodingdefault}{\sfdefault}{bx}{n}

\usepackage{amsmath}
\usepackage{graphicx}
\usepackage{xcolor}
\usepackage{listings}
\usepackage{framed}
\usepackage{url}
\usepackage{hyperref}
\title{A Modular LLM Framework for Explainable Price Outlier Detection}

\author{Shadi Sartipi,~~ John Wu\thanks{This work was done as part of an internship at Amazon.},~~ Sina Ghotbi,~~ Nikhita Vedula, ~~ Shervin Malmasi \\
Amazon.com, Inc.\\
\texttt{\{ssartipi,ghotbis,veduln,malmasi\}@amazon.com} \\
}

\iclrfinalcopy %
\begin{document}

\maketitle

\begin{abstract}
Detecting \textit{product price outliers} is important for retail and e‑commerce stores as erroneous or unexpectedly high prices adversely affect competitiveness, revenue, and consumer trust. Classical techniques offer simple thresholds while ignoring the rich semantic relationships among product attributes. We propose an agentic Large Language Model (LLM) framework that treats outlier price flagging as a reasoning task grounded in related product detection and comparison. The system processes the prices of target products in three stages: (i) relevance classification selects price‑relevant similar products using product descriptions and attributes; (ii) relative utility assessment evaluates the target product against each similar product along price influencing dimensions (e.g., brand, size, features); (iii) reasoning‑based decision aggregates these justifications into an explainable price outlier judgment. The framework attains over 75\% agreement with human auditors on a test dataset, and outperforms zero‑shot and retrieval based LLM techniques. Ablation studies show the sensitivity of the method to key hyper-parameters and testify on its flexibility to be applied to cases with different accuracy requirement and auditor agreements. 
\end{abstract}

\section{Introduction}
Sharp pricing is essential for fair competition and consumer trust in large-scale e-commerce, and has motivated recent work in accurate price distribution estimation \citep{vedula-etal-2025-quantile}.
Pricing errors can lead to significant financial losses, customer service contacts, and negative publicity \citep{ramakrishnan2019pricing}. 
Anomaly detection methods from the broader literature provide foundational techniques including statistical approaches \citep{rousseeuw2011robust, chandola2009anomaly}, density estimation methods \citep{dudik2007hierarchical}, support vector-based novelty detection \citep{scholkopf1999support}, and tree-based isolation methods \citep{liu2008isolation}. However, when applied to pricing contexts, these approaches face critical limitations like relying on heuristic thresholds that provide no rationale for why a price is considered an outlier, making them unsuitable for audit-critical retail systems \citep{sakthivanitha2025real, olteanu2023meta}. 

Existing approaches for identifying comparable products in e-commerce face several challenges. While recent advances in product description generation and question-answering systems have improved customer experience \citep{tangarajan2025contextually, wang2024leveraging}, embedding-based retrieval and product matching systems still struggle to capture price-relevant semantic relationships \citep{BERT_product_matching, ristoski2018machine}. These challenges include data heterogeneity, varying levels of data quality, and difficulty in representing products with missing specifications \citep{BERT_product_matching, ristoski2018machine}. Products retrieved based on similarity may differ in key price-influencing attributes such as pack size, brand tier, or functionality, introducing noise into pricing decisions. To overcome this, human auditors first identify truly comparable substitutes, then weigh value dimensions like brand, quantity, and feature richness before deciding whether a deviation is justified \citep{ramakrishnan2019pricing}. Recent advances in LLMs offer a promising approach to automate this human-like reasoning process.

LLMs demonstrate strong contextual understanding, utilizing multiple tools and chain-of-thought reasoning \citep{yao2022react, shinn2023reflexion,xi2025rise}. Retrieval-augmented and graph-augmented frameworks further enhance multi-entity grounding \citep{lin2025srag, edge2024local}, while enterprise-grade models such as Claude Sonnet and OpenAI o3 emphasize cost efficiency and reliability for real-world deployment \citep{anthropic2024introducing, amazon2024nova, openai2025o3}. The capabilities of these approaches motivate treating price anomaly or outlier detection as a structured reasoning task that mirrors human audit logic in selecting semantically relevant products, assessing relative utility across product-value dimensions, and generating an interpretable rationale. To our knowledge, no existing automated system emulates this multi-step reasoning process for price anomaly detection.

In this study, we propose an agentic\footnote{We use ``agentic'' to refer to multi-step LLM architectures with specialized reasoning modules that decompose complex objectives into interpretable sub-tasks, consistent with recent literature \citep{yao2022react, xi2025rise}. Our framework involves sequential coordination between modules rather than dynamic inter-agent communication.} LLM framework for explainable price-outlier detection focused on too high prices. Our framework consists of three specialized agents chaining processing in three steps. They explicitly model the factors human auditors would consider. (i) Step one performs \emph{Relevance Classification} by using product details and structured attributes to identify product candidates similar to or substitutes for the target product, which we refer to henceforth as \textit{neighbors}. (ii) Step two conducts \emph{Relative Utility Assessment}, reasoning about key price-influencing attributes such as brand, unit size, and other feature values. (iii) Step three aggregates the above evidence into a \emph{Reasoning-Based Decision} with a justification for outlier understanding. We evaluate our proposed framework on products spanning several categories which achieves 76.3\% agreement with human expert annotations and reduces the false-discovery rate to 7.8\%. Moreover, detailed ablation studies on challenging test cases show that the count of neighbor products, price padding thresholds, and product attribute selection process can affect performance. Beyond pricing, this formulation can potentially extend to adjacent applications such as product matching and fraud detection, contributing to the broader adoption of trustworthy AI.
\begin{figure*}[t]
    \centering
    \includegraphics[width=0.8\linewidth]{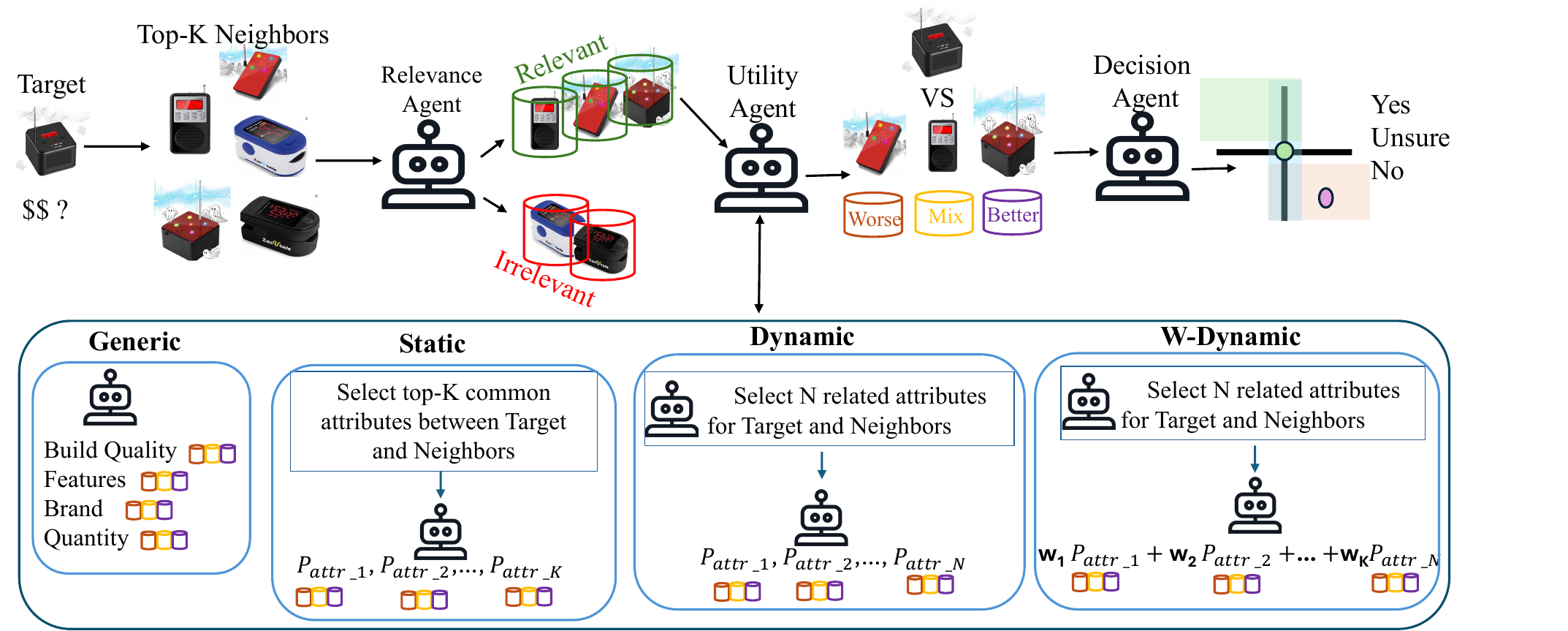}
    \caption{Overview of our proposed agentic framework. We investigate four attribute selection configurations with the Utility Agent (Generic, Static, Dynamic, or W-Dynamic), detailed further in Section \ref{ss:Relative_Utility_Assessment}.}
    
    \label{fig:main-flowchart}
\end{figure*}

\textbf{Application contribution}: Price outlier detection directly impacts revenue and consumer trust at scale. Our framework achieves a high agreement with human auditors while providing transparent justifications, a capability absent from existing statistical methods (e.g., IQR, Z-score) and rule-based systems. The explainability enables human auditors to verify and override decisions efficiently.

\section{Related Work}
\subsection{Price Anomaly Detection}
Classical anomaly detection methods rely on univariate thresholds \citep{olteanu2023meta} and multivariate techniques such as Isolation Forest, Local Outlier Factor, and deep autoencoders \citep{samariya2023comprehensive}. While effective at capturing statistical irregularities, these approaches remain largely black-box with limited interpretability, unable to explain which neighbors are compared, and why the data density differs. Industrial pricing systems have attempted to address this through retrieving substitute and similar products combined with rule-based scoring \citep{ramakrishnan2019pricing, sarpal2023marketplace}, yet these solutions still lack human-readable rationales for detected anomalies and heavily rely on hard-coded rules. 
In \citep{sarpal2023marketplace}, the authors developed a high price anomaly detection system for a growing marketplace platform, leveraging three modular components to engineer features, detect feature-level anomalies, and build reliable item price estimates. Recent surveys on out-of-distribution detection and dynamic graph anomalies \citep{yang2024generalized, ekle2024anomaly} highlight the growing need for contextual, structure-aware detectors. These challenges are particularly relevant for e-commerce systems with non-stationary product catalogs and evolving catalog dynamics. We applied this motivation by integrating semantic reasoning with natural language explanations directly into our anomaly detection pipeline, enabling accurate detection and interpretable justification of pricing anomalies.

\subsection{Agentic LLM Architectures}
Instruction-tuned LLMs exhibit strong chain-of-thought reasoning and tool-use capabilities \citep{yao2022react, shinn2023reflexion, xi2025rise}.
These agentic architectures decompose complex objectives into interpretable sub-tasks, enabling multi-step problem solving with intermediate reasoning steps for audit and control.
Recent surveys highlight their application across planning, decision-making, and interactive environments \citep{xi2025rise}, though adoption in enterprise backend systems like pricing remains limited. Recent work has also explored LLMs for anomaly and outlier detection across various domains \citep{su2024large}.

\subsection{RAG for Structured Data}
Retrieval-augmented Generation (RAG) improves LLMs' factual grounding by incorporating external knowledge \citep{asai2023self}.
Graph-augmented retrieval extends this to entity-aware summarization over structured, relational data \citep{edge2024local}.
In e-commerce, product-aware RAG pipelines leverage catalog metadata to enhance search, question answering, and product understanding \citep{sun2024productrag, hosseini2024llmaj, wang2024leveraging, tangarajan2025contextually}.
LLM-based ranking frameworks further enable listwise comparison and re-ranking over retrieved candidates \citep{ma2023zero, gao2024llm4rerank}.

Our work introduces an agentic LLM framework that reframes price outlier detection as a structured reasoning task. Our three-step workflow of relevance classification, relative utility assessment and reasoning-based decision generation, grounds each step in explicit product semantics (titles, attributes, unit prices) to reduce the need and complexity of expensive human audits. Unlike post-hoc LLM explanations that may not reflect true decision processes \citep{madsen2024faithful, huang2023canllmsexplain}, our approach generates verifiable reasoning chains anchored in structured product data, ensuring transparency and auditability at scale, essential for enterprise pricing. Our key contributions are: (i) a novel methodological framework that decomposes price outlier detection into interpretable reasoning sub-tasks, (ii) demonstration of practical applicability in supporting human audit workflows with explainable decisions, and (iii) comprehensive ablation studies revealing key trade-offs in system configuration.

\section{Method}
\label{sec:method}
In this study, given a target product, we seek to identify if it has an \textit{anomalously high price}, given the observed prices of comparable products. Our ground truth consists of human audit decisions on the product prices, given the context of similar or comparable neighbor products, and their corresponding median prices 
across multiple other sellers. Given the cross-seller nature of these neighbor prices, we posit that they are credible for informing about the target product's price quality. Since these neighbor products are not identical to the target, we incorporate their pricing value differentiators into the problem solution. Finally, we aggregate these price signals into a final classification between anomalously high price, ambiguous (unsure), and not anomalously high price categories. Our proposed agentic framework models this problem as a chain of reasoning steps across three sequentially coordinated agents, explained below.

\subsection{Candidate Neighbor Product Generation}
Price-relevant neighbor products are ones that a customer would take their prices into account when judging the price of a target product. These products are the ones with similar functionality, like color and size variations of the same brand. Moreover, the products should have reliable pricing data like multiple sellers offer. We rely on proprietary product embeddings trained using contrastive learning on product titles and images, where text and image representations are projected into a shared latent space (similar to CLIP-style architectures). Then we use nearest neighbor embedding-based cosine similarity search to identify the top-$k$ \textit{candidate neighbors} for a given target product. 
\subsection{Relevance Classification Agent}
The first agent in our proposed framework is the LLM-based \textit{Relevance Agent}, which determines whether each candidate neighbor is relevant to the target product. Let $\mathcal{A} = \{A_1, A_2, ..., A_k\}$ represent the set of candidate neighbor products and $T$ represent the target product. The agent decides that a neighbor $A_i$ is price-relevant when its answer to any of the following questions is ``Yes'' (disjunctive logic, as each question captures a different aspect of price-relevance), and also provides an explanation for its decision: ``Would customers compare prices between the target and this product while buying the target?'', ``Is this product similar to the target product?'', and ``Can this product be an alternative serving the same use case?''. These questions were formulated based on how human pricing auditors assess comparability, derived from domain expert interviews and literature on consumer price comparison behavior \citep{ramakrishnan2019pricing}. See Appendix~\ref{ss:prompt_template} for prompt details.

The Relevance Agent evaluates the top-k neighbors obtained via the similarity search mentioned earlier. 
Increasing the number of neighbors $k$ provides the LLM more product context, but also adds to the computation overhead and could cause confusion for the downstream agents. Therefore, it is important to experimentally find the right balance for this hyper-parameter, as we illustrate in Section~\ref{ss:padding_neighbors}. 

\subsection{Utility Assessment Agent}
\label{ss:Relative_Utility_Assessment}
The second agent in our framework is the \textit{Utility Agent}, which compares each neighbor product to the target based on key attributes that are likely to influence price. The selection of these attributes is critical for this framework. We introduce and evaluate four configurations for attribute selection (shown in Figure~\ref{fig:main-flowchart}).\\

(i) \textbf{Generic}: Fixed, general price driving criteria applicable to all product categories, namely, build quality, features, brand reputation, and quantity. See example prompt template in Appendix \ref{ss:prompt_template}.

(ii) \textbf{Static category-specific}: For each product category, we independently generate the top key price-influencing attributes (e.g. wattage, ingredients) using a zero shot foundational LLM. 

(iii) \textbf{Dynamic product pair specific}: For each target-neighbor product pair and available product details, have the Utility Agent itself generate the top $N$ pricing influencing attributes.

(iv) \textbf{Weighted Dynamic (W-Dynamic) product pair specific}: For each target-neighbor pair and available product details, have the Utility Agent dynamically generate the top $N$ price driving attributes and also assign a relative importance weight to each attribute.

The Utility Agent uses its judgment to compare the target product to each candidate neighbor based on these attributes. 
On each attribute dimension, the agent decides if the neighbor is better (+1), mixed (0), or worse (-1) than the target product. Then it sums, or sums and weights the score on all attributes. That net utility along with the price point of the neighbor product provides a vote for the Decision Agent to make the final call on target price outlier status.

\begin{figure}[t]
\centering
\begin{minipage}{0.7\linewidth}
    \centering
    \includegraphics[width=0.48\linewidth]{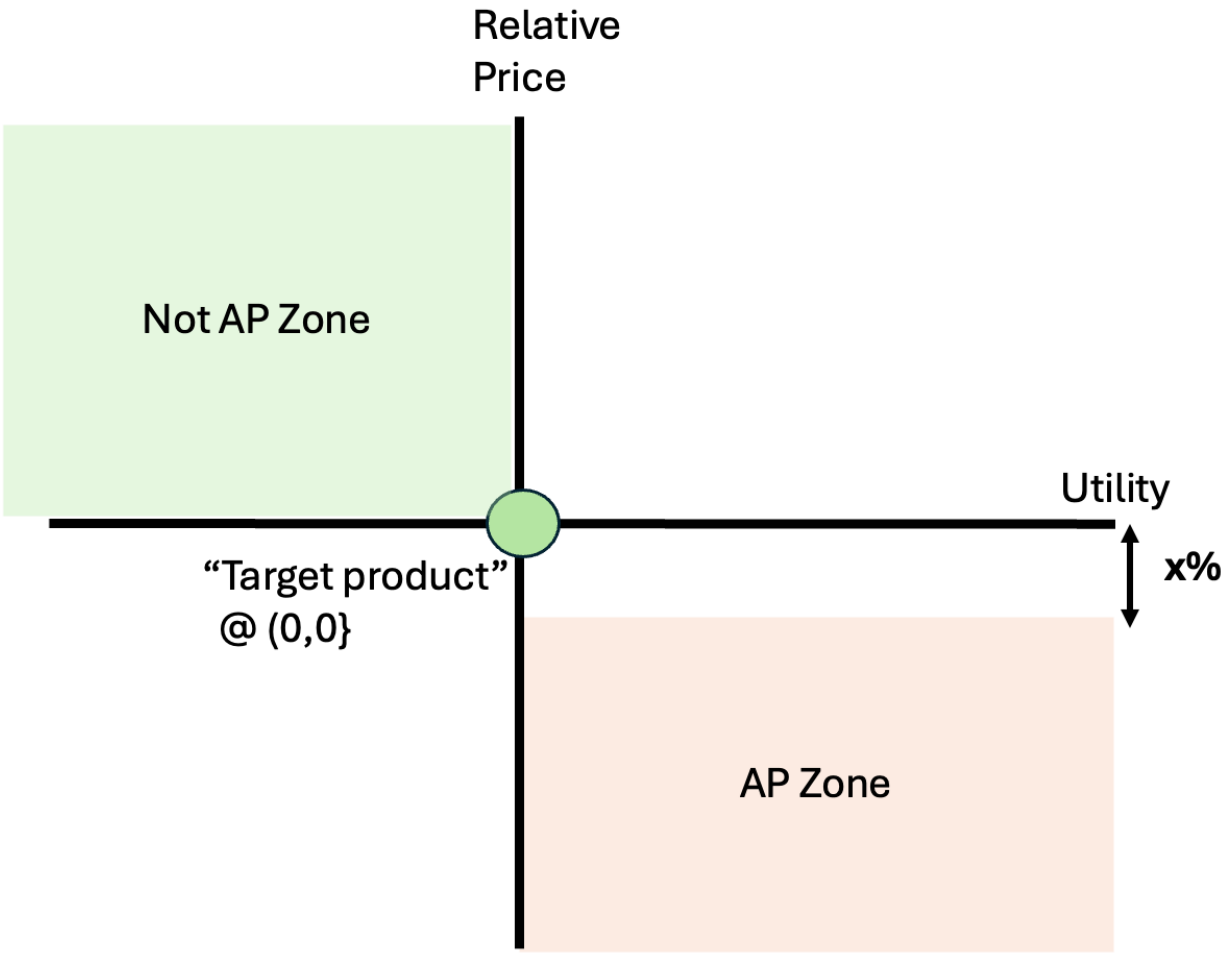}
    \hfill
    \includegraphics[width=0.48\linewidth]{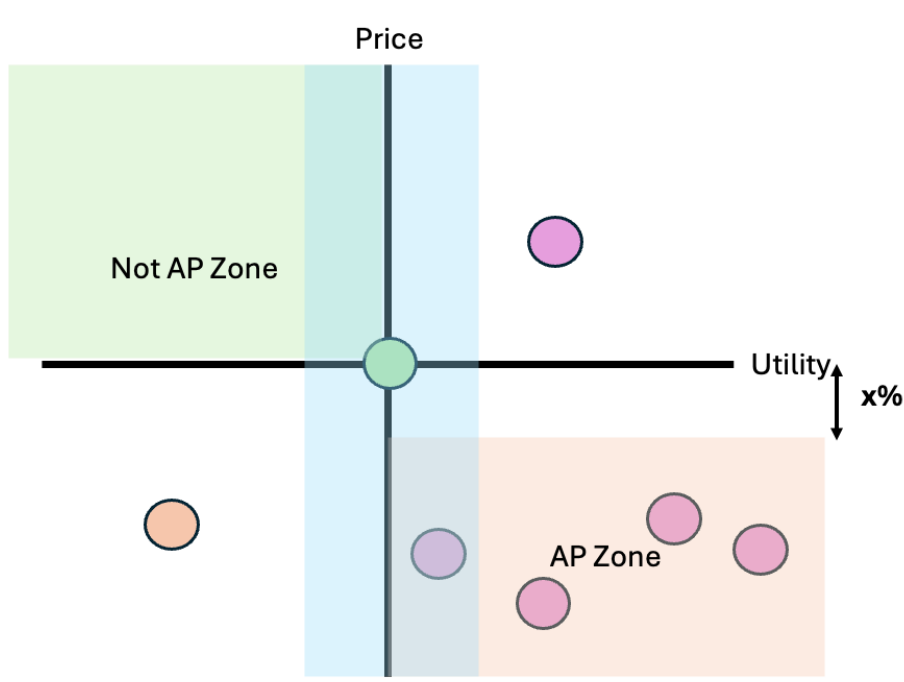}
\end{minipage}
\vspace{0.5em}
\begin{minipage}{0.8\linewidth}
    \centering
    (a)
\end{minipage}
\vspace{1em}
\begin{minipage}{0.7\linewidth}
    \centering
    \includegraphics[width=0.48\linewidth]{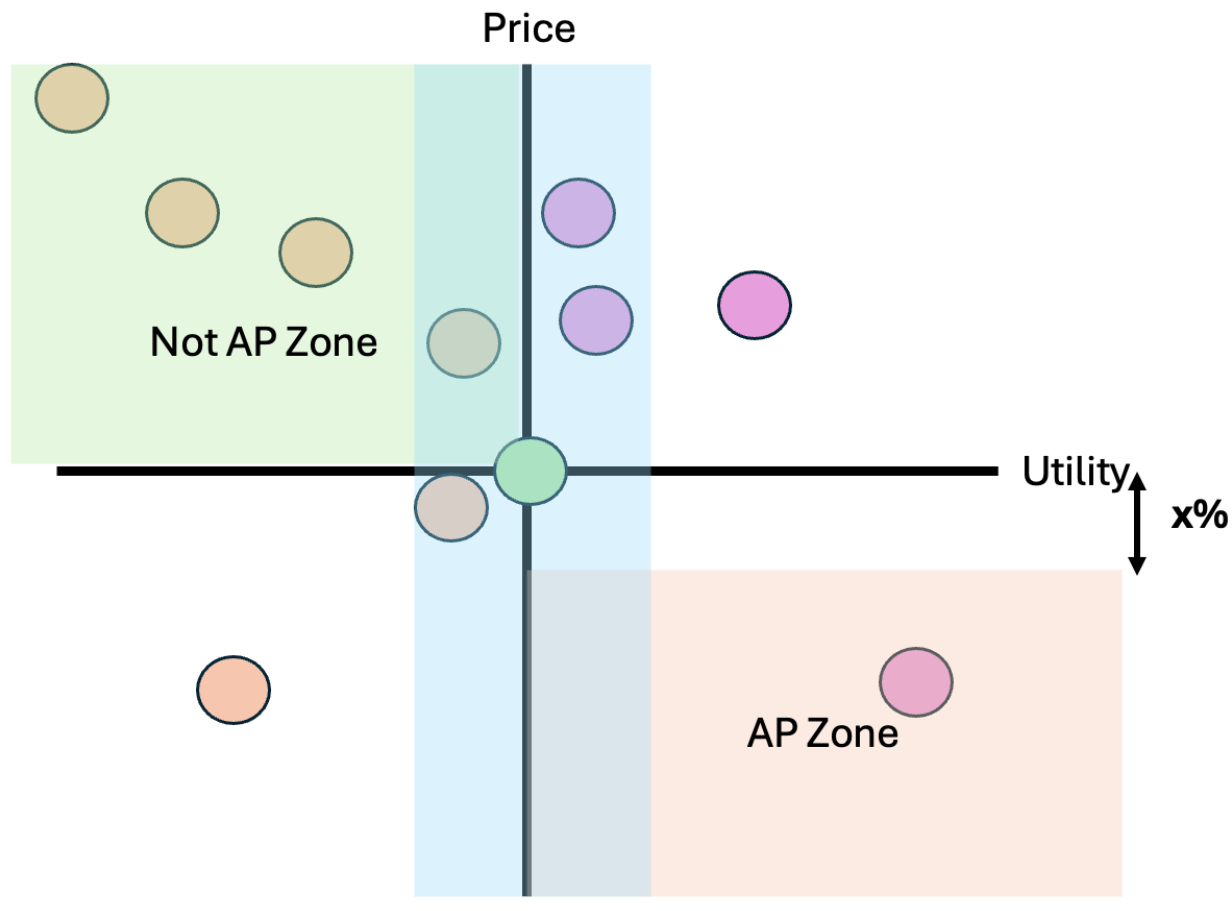}
    \hfill
    \includegraphics[width=0.48\linewidth]{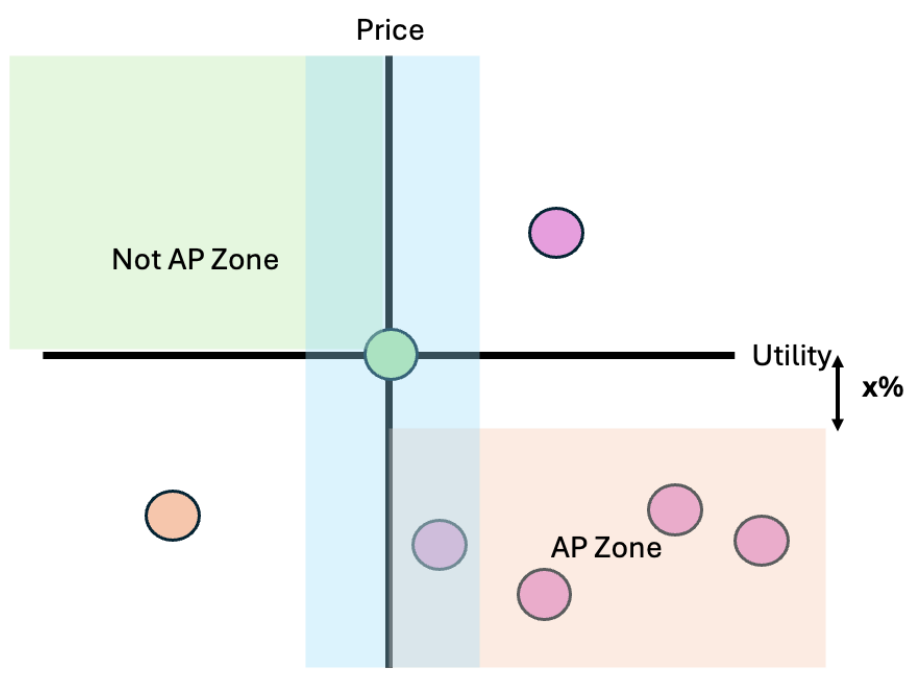}
\end{minipage}
\vspace{0.5em}
\begin{minipage}{0.8\linewidth}
    \centering
    (b)
\end{minipage}
\caption{(a) The target 4-quadrant framework showing (\textbf{left}) the informative zones and (\textbf{right}) relevant products shown on uninformative zones along with utility trade-off padding. (b) The sample of the decision showing the (\textbf{left}) correct price, and (\textbf{right}) outlier price.}
\label{fig:quadrants}
\end{figure}

\subsection{Reasoning-Based Price Decision Agent}
A final pricing decision agent determines whether the target product's price is an outlier by assessing the evidence from the prior steps. We propose to represent the information out of the first two agents on four quadrants, delineated by relative price and utility axes as illustrated in Figure~\ref{fig:quadrants}. The target product is placed at the origin and neighbors are positioned relative to that. The candidate neighbors filtered by the first agent and with higher (lower) utility than the target, as judged by the second agent, will be on the right (left) quadrants. Neighbors with price points above the target will be on the top quadrants. Price anomaly detection for the target product is thus guided by point positions in the quadrants. 

The top-right quadrant represents relevant products with higher utility and higher price (likely better and pricier). These do not provide direct evidence on price quality of the target because such higher value products are expected to be more expensive than the target. On the other hand, the bottom-right quadrant is where we see higher utility and lower priced relevant (likely better and cheaper) products. Data points in that quadrant, called the \textit{Anomaly Price (AP) zone}, support the hypothesis that the target product is priced too high. When products of similar or better value are priced cheaper than the target product by a considerable margin, the target product is likely to have anomalously high price. Inversely, top-left (worse and pricier) neighbors provide evidence against an anomalous high price. Bottom-left (worse and cheaper) neighbors, like top-right quadrant, do not provide direct evidence on anomalously high price of target product.

In addition, we propose a \textit{padded region of trade-offs} to control for noisiness or uncertainty of the utility comparison process (Figure~\ref{fig:quadrants} (a) right). We assume that net utility should deviate beyond the padded region for a relevant product to be considered better or worse than the target, otherwise it is in the trade-off zone and practically should be considered similar utility. Similarly on the price axis, we define \textit{price padding} as the minimum percentage of price difference between the target and neighbor that indicates an anomalous price. This is a hyper parameter that controls the sensitivity of the process to outlier price detection. A larger padding reduces the false positives at the cost of more false negatives. We empirically study the impact of this hyperparameter in Section \ref{ss:padding_neighbors}.
Finally, given the data points in the four quadrants, the Decision Agent can use two strategies to decide if a target product has an outlier price.

(i) \textbf{Worse-Pricier Veto}: Presence of a single relevant product in the NOT AP Zone indicates that the target product price is not anomalous. This can be used when we highly trust the neighbor products, and seeing a single case justifies that the target price is not anomalously high.

(ii) \textbf{Quadrant Voting}: Presence of equal or more products on the AP Zone than NOT AP Zone indicates that the target price is anomalous. This can be used when we have moderate confidence in all neighbors and want to rely on a collective agreement among them.

This specification also serves as a categorical hyper-parameter that we report experimental results on in Section \ref{sec:eval}. We acknowledge that these settings and alternatives are not the only possibilities and depending on the application other arrangements should be experimented with.

\begin{table*}[t!]
\centering
\scriptsize
\setlength{\tabcolsep}{4pt} %
\begin{tabular}{lcccccccc}
\hline
\textbf{Approach} & \textbf{SS F1}$\uparrow$ & \textbf{SS Pre.}$\uparrow$ & \textbf{SS Rec.}$\uparrow$& \textbf{Agreement}$\uparrow$ &\textbf{EC F1}$\uparrow$ & \textbf{EC Pre.}$\uparrow$ & \textbf{EC Rec.}$\uparrow$& \textbf{Outlier Rate} \\
\hline
\multicolumn{9}{l}{Baseline Comparison}\\
\hline
Zero Shot & 0.43 & 0.50 & 0.38 & 68\%&0.50& 0.33& 1.0 & 15.1\%\\
RAG & 0.42 & 0.31 & \bf{0.63} & 23.0\%&0.29& 0.17& 1.0&40.9\% \\
Agentic & \bf{0.55}& \bf{1.00} & 0.39 & \bf{76.3\%} &\bf{0.50}& \bf{0.33}& 1.0&\bf{7.8\%} \\
\hline
\multicolumn{9}{l}{Agentic - Price Paddings (@7 Neighbors)}\\
\hline
30\% & \bf{0.71} & 0.67 & 0.75 & 52.6\%&0.29& 0.17& 1.0 & 14.9\%\\
50\% & 0.55 & \bf{1.0} & 0.38 & 76.3\%&0.50& 0.33& 1.0&7.8\% \\
75\% & 0.0& 0.0 & 0.0 & \bf{94.6\%} &\bf{0.67}& \bf{0.50}& 1.0&\bf{2.4\%} \\
LLM-Based padding & 0.67& 0.54 & \bf{0.88} & 42.0\% &0.50& 0.33& 1.0&10.4\% \\
\hline
\multicolumn{9}{l}{Agentic - Number of Neighbors (@50\% padding)}\\
\hline
3 & 0.20 & 0.50 & 0.13 & 72.5\%&0.50& 0.33& 1.0 & 9.6\%\\
7 & \bf{0.55} & \bf{1.00} & \bf{0.38} & \bf{76.3\%}&0.50& 0.33& 1.0&7.8\% \\
12 & 0.33& 0.40 & 0.29 & 71.0\% &0.50& 0.33& 1.0&\bf{6.9\%} \\
\hline
\multicolumn{9}{l}{Agentic - Price Influencing Attribute Configuration (@50\% padding and 7 Neighbors)}\\
\hline
Generic & \bf{0.55} & \bf{1.0} & \bf{0.38} & 76.3\%&\bf{0.50}& \bf{0.33}& 1.0 & 7.8\%\\
Static cat-spec & 0.33 & 0.50 & 0.25 & 67.6\%&0.33& 0.20& 1.0&9.6\% \\
Dynamic prod-pair spec& 0.31& 0.40 & 0.25 & 59.5\% &0.50& 0.33& 1.0&10.2\% \\
W-Dynamic prod-pair spec& 0.20& 0.50 & 0.12 & \bf{81.6\%} &0.50& 0.33& 1.0&\bf{6.4\%} \\
\hline
\multicolumn{9}{l}{Agentic - Decision Strategy (@50\% padding, 7 Neighbors, and Static Cat-spec)}\\
\hline
Quadrant Voting& 0.50 & 0.42 & \bf{0.63} & 43.2\%&0.50& 0.33& 1.0 & \bf{5.6\%}\\
Worse-pricier veto & \bf{0.55} & \bf{1.0} & 0.38 & \bf{76.3\%}&0.50& 0.33& 1.0&7.8\% \\
\hline
\end{tabular}
\caption{Comparing our agentic approach with baselines and hyper-parameter variations on the silver set (SS), one-sided annotated set (measured by agreement score), edge cases (EC), and outlier rate on the unannotated set. `Pre.' and `Rec.' stand for Precision and Recall respectively.}
\label{tab:price_padding}
\end{table*}

\section{Experiments and Results}
\label{sec:eval}
\subsection{Dataset}
We construct a dataset by crawling multiple popular online shopping stores to obtain the product names, details and their corresponding prices across multiple sellers. We then split our dataset into four subsets. The first is a balanced annotated set (silver set or SS) where we have 40 human annotated products across product categories with half of them labeled as having an anomalously high price. Up to 3 annotators per product are used to ensure a majority agreement. Our annotators are domain expert pricing auditors. We focused our evaluation on products where at least 2 annotators agreed on the labeling decision, yielding a silver-standard annotation set given the inherently subjective nature of atypical pricing judgments.
The second subset is a one-sided annotated set containing 185 products representing difficult cases (falsely flagged as high price by internal systems) that are all labeled as \textit{not} anomalous by human annotators.  
The third is a subset of edge cases (EC) consisting of 12 products with unique characteristics (e.g. huge pack-size, portable home, old iPhone versions) which are difficult to reliably price and assess for outliers. 
The fourth subset is an unannotated set of 5400 high traffic products across product categories (e.g, apparel, home goods, and personal care items), to measure the overall rate of outlier flagging of our agentic framework. This is a guardrail metric that is expected to be below 10\% and ideally closer to 5\%, based on general expectations of price quality. We also use this subset to measure our agentic framework's scaling in time and cost, shown in Appendix \ref{ss:compute_and_scaling}.

In summary, our test sets contain the Silver Set (40 products with 50\% outlier rate), the one-sided annotated set (185 products with 0\% outlier rate), Edge Cases (12 products with 25\% outlier rate) and the unannotated set which comprises 5400 products.

\subsection{Results}
We use Claude 3.5 Sonnet as the LLM powering our agents and evaluate the performance on the four mentioned testsets on eight metrics in Table~\ref{tab:price_padding}. For the Silver Set (SS) and Edge cases (EC) sets, we report precision, recall and F1-score. For the one-sided annotated set, we report the agreement percentage between the agentic framework and annotators on flagging challenging non-outlier cases. For the unannotated, set we report the predicted outlier rate percentage. We compare our agentic approach with other LLM-based approaches, and also investigate variations of price-padding, number of neighbors, price driving attribute types and final decision strategies outlined in Section \ref{sec:method}.

\subsubsection{Baseline Comparisons}

Table~\ref{tab:price_padding} represents the performance comparison of our proposed agentic framework against (i) a Direct Retrieval (RAG) baseline that just retrieves and uses the top k embedding similarity based neighbors to make a price outlier judgment without structured reasoning; and (ii) a Single-Call LLM (Zero Shot) baseline that only uses the target product's details and price to make a price outlier judgment without any neighbor context. These baselines represent common industry approaches: the RAG baseline reflects typical embedding-based product comparison systems, while the Single-Call baseline tests whether an LLM can detect outliers from product information alone. Traditional statistical methods (IQR, Z-score) are not directly comparable as they require historical price distributions and provide no explainability.

Comparing to these two baseline methods, our agentic approach (at 50\% price padding, 7 neighbors, generic attributes, and worse-pricier veto) achieves the highest SS F1 score of $0.55$ and precision of $1.00$, indicating that the identified outlier prices are consistently correct according to the SS human annotations. Moreover, our approach achieves over 76\% agreement rate with the one-sided non-anomalous annotated data subset, and has a reasonable outlier rate less than 8\%. Given the inherent subjectivity of price quality assessment (human annotators themselves agree at $\kappa = 0.72$), our 76.3\% system-human agreement demonstrates strong alignment with expert judgment. Additionally, the one-sided annotated set represents ``difficult cases'' where products were flagged by internal systems but overruled by human auditors - achieving 76.3\% agreement on these challenging edge cases demonstrates robustness. These results indicate the superiority of our agentic approach over the baselines. %

\subsubsection{Impact of Price Padding and Neighbors}
\label{ss:padding_neighbors}
Next, we investigate our framework's performance on price padding variations. A 30\% price padding achieves the highest SS F1 score of $0.71$ with balanced precision and recall, while 50\% padding maximizes precision at the cost of recall. Allowing the LLM agent to decide the price padding threshold results in the highest recall (0.88) with moderate precision (0.54).
Increasing price padding reduces the outlier rate, but decreases our agreement with SS. Increasing the padding also causes the agent to struggle to detect EC outliers. These mixed results provide evidence on price padding being a key knob  controlling the sensitivity, and the need to fine-tune it given the task of interest. For the rest of the experiments, we set the padding to $50\%$ which serves as a middle-ground between F1 score and agreement on the one-sided challenging test set, while keeping outlier rate below 10\%.

The next part of Table~\ref{tab:price_padding} shows the effect of varying the number of neighbors.
7 neighbors seems to hit a sweet spot for the SS-F1 and the one-sided test subset agreement score, while keeping the outlier rate low (the EC set is not sensitive to this dimension). This finding indicates that too few neighbors can lose context while too many neighbors can overload the agent with too much context.

\subsubsection{Impact of Price-Influencing Attributes and Decision Strategy}
We investigate the performance of the four different price-influencing attribute selection configurations detailed in Section~\ref{ss:Relative_Utility_Assessment}. For the Dynamic product-pair configuration, we choose the minimum number of neighbors as $3$, and the maximum number is decided by the LLM agent depending on the target product and its neighbors' attributes.
The W-Dynamic configuration achieves the lowest outlier rate of 6.4\% and highest agreement with annotations (81.6\%), but at the cost of significantly reduced SS F1 ($0.20$). The Generic configuration provides the best overall balance with the highest SS F1 of $0.55$, perfect precision ($1.0$), and strong agreement rate of 76.3\%. The Static category-specific and Dynamic product-pair specific configurations show low SS F1 and agreement scores. These results surface a key trade-off between consistent high level prompting vs adaptive attribute selection cases or overloading the LLM with judgment decisions. The Generic approach's relative success (best in F1 scores and a close second on agreement) demonstrates that effective agentic systems benefit from a clear structure with an optimal level of autonomy delegated to the LLM.

Finally, we evaluate the two decision strategies for determining price anomalies (Table~\ref{tab:price_padding}, bottom section). Worse-Pricier Veto achieves high precision ($1.0$) and SS F1 of $0.55$ with moderate recall ($0.38$). Quadrant Voting significantly improves recall from $0.38$ to $0.63$, but reduces precision from $1.0$ to $0.42$. The human agreement on the one-sided test drastically improves from 43\% to 76\% with the Worse-Pricier Veto.
These results demonstrate that the choice of decision strategy affects the aggressiveness of outlier flagging. Worse-Pricier Veto prioritizes precision, minimizing false positives at the expense of lower recall, while Quadrant Voting increases recall but introduces more false positives. This tunability allows the framework to be adapted to different application contexts: high-precision configurations suit consumer-facing systems where false alarms erode trust, while high-recall configurations are appropriate for internal auditing where missing anomalies is more costly. Interestingly, this result supports the hypothesis that human auditors mostly follow the worse-pricier veto approach in their decision making, which clearly imposes a lower cognitive load.

The primary downstream application of our framework is supporting human audit workflows in e-commerce pricing systems. Flagged products are reviewed by auditors who can inspect the LLM-generated explanations, significantly reducing audit time compared to black-box systems. We ensure annotation quality through domain expert pricing auditors with majority agreement among up to 3 annotators per product. Our evaluation spans multiple complementary test sets (standard, difficult, edge cases) for comprehensive performance assessment. The explainability enables auditors to quickly verify decisions by examining which neighbors were compared and why the utility comparison led to the outlier flag. The outlier rate metric (7.8\%) serves as a guardrail aligned with business expectations that fewer than 10\% of products should be flagged for review, ensuring the system does not overwhelm human reviewers.

\section{Conclusion}
This paper presents an agentic LLM framework for explainable price outlier detection in e-commerce, that treats price flagging as a reasoning task grounded in product semantics. Our three-stage approach (relevance classification, utility comparison, and reasoning-based decision) achieves over 75\% agreement with human auditors on a challenging test set which demonstrates that agentic, explanation-driven LLMs can deliver transparent, category-aware pricing at a practical cost. 

\section{Limitations}
Our implementation represents an initial exploration of the agentic framework's capabilities rather than a fully optimized solution. Several areas warrant further investigation. First, the utility quantification can be made more unbiased and representative by running an aggregated investigation over several different customer personas represented by LLMs. Second, improving the current attribute selection process may improve system performance, particularly in dynamic configurations. Third, our framework assumes that product metadata (attributes) is available and reasonably accurate; in practice, e-commerce data is often incomplete or noisy, which may affect utility comparisons. Our framework handles missing attributes by comparing only available information, but systematic data quality issues could impact performance. We leave a detailed data quality analysis and potential mitigation strategies (e.g., confidence weighting based on attribute completeness) for future work. Finally, comprehensive evaluation of such solutions remains challenging due to the inherent contextuality and subjectivity of price-quality ground truth, resulting in limited high-quality evaluation data. We note that what constitutes an ``anomalously high'' price is inherently subjective and depends on consumer perception, market context, and individual judgment, this subjectivity is reflected in our inter-annotator agreement ($\kappa = 0.72$) and represents a fundamental challenge for any evaluation methodology in this domain. These limitations highlight important directions for future research in this domain.

\bibliography{iclr2026_conference}
\bibliographystyle{iclr2026_conference}

\appendix
\section{Appendix}
\subsection{Prompt Templates and Agent Instructions}
\label{ss:prompt_template}
Here are the prompts that the agents use for different sections. 

Note: The template syntax (e.g., \texttt{\{len(static-attributes)\}}, \texttt{\{attribute-list\}}) represents placeholder variables that are dynamically filled at runtime. This is pseudo-code notation, not actual Python execution within the prompt.

\begin{framed}
\tiny
\begin{verbatim}
"""You are a pricing relevance expert. Determine if a neighbor product 
    is relevant for analyzing the target product's pricing strategy and market positioning.
    
    Categories:
    - "Relevant": Products that customers would directly compare prices for when making 
      purchasing decisions. These are direct substitutes, very similar products, or products 
      that serve the same primary use case and would influence purchasing decisions.
    - "Irrelevant": Products that provide no meaningful pricing context or comparison value.
      This includes products in different categories, different price tiers, or serving 
      different primary needs.
    
    Consider these factors:
    - Would customers compare prices between these products when making a purchase decision?
    - Do they serve the same primary use case or need?
    - Are they direct substitutes or close alternatives?
    - Would one product's pricing directly influence the other's market positioning?
    
    Use a stricter standard than typical relevance assessment - only classify as "Relevant" 
    if the products are direct competitors that customers would actively compare.
    
    Format your response as a JSON object with two fields:
    1. 'explanation': Your detailed reasoning for the pricing relevance classification 
    (limit to 50 words maximum)
    2. 'relevance': Must be exactly "Relevant" or "Irrelevant"
    
    Example response:
    {
        "explanation": "Both are premium wireless headphones in the same price range that customers would 
        directly compare when choosing between brands for identical use cases.",
        "relevance": "Relevant"
    }"""
\end{verbatim}
\end{framed}

\begin{framed}
\tiny
\begin{verbatim}
"""You are a product comparison expert using STATIC attribute mode.

CRITICAL: You MUST use these EXACT \{len(static-attributes)\} attributes - NO selection or substitution 
allowed:

\{attribute-list\}

Your task is to:
1. Compare the two products using ALL \{len(static-attributes)\} attributes listed above
2. Assign utility scores (1-3) based on importance for this specific comparison
3. Include Brand as a critical comparison with utility = 3 (critical differentiator)
4. Include Quantity with detailed composition breakdown and utility = 3
5. Provide detailed analysis for each attribute

For each attribute, specify if the Neighbor PRODUCT is "better", "worse", "same", or "mixed" vs the BASE 
PRODUCT.

Consider target demographics: bulk buyers, individual consumers, commercial users.

STATIC MODE RULES:
- You MUST use ALL \{len(static-attributes)\} attributes listed above - no exceptions
- NO selection or generation of different attributes allowed
- Focus on functional attributes that impact product functionality and user experience
- Use the exact attribute names as provided above


MANDATORY REQUIREMENTS:
- Use ALL \{len(static-attributes)\} attributes listed above (no selection allowed)
- Brand must ALWAYS be included as a attribute with utility = 3 (critical differentiator)
- Quantity must ALWAYS be included with detailed composition breakdown and utility = 3

CRITICAL: Return ONLY pure JSON - no markdown code blocks, no explanations, no additional text.
\end{verbatim}
\end{framed}

The following shows the Dynamic mode prompt template:

\begin{framed}
\tiny
\begin{verbatim}
"""You are a product comparison expert with intelligent attribute selection capabilities.

Your task is to:
1. ANALYZE both product descriptions to identify which attributes are mentioned, implied, or can be 
reasonably inferred
2. MATCH available attributes to the product content - only select attributes where both products 
have relevant information
3. SELECT exactly \{self.top$_{N-attributes}$\} attributes that are most relevant and differentiating 
for these specific products
4. Compare the products using these selected attributes with utility scoring
5. Include Quantity as a critical attribute with detailed composition breakdown for all comparisons
6. Provide detailed analysis for each attribute

For each attribute, specify if the Neighbor PRODUCT is "better", "worse", "same", or "mixed" vs the BASE 
PRODUCT.
Also assign utility scores (1-3) based on importance for this specific comparison.
"""
\end{verbatim}
\end{framed}

\begin{framed}
\tiny
\begin{verbatim}
"""You are a pricing analysis expert. Determine if the target product is anomalous priced.
SIMPLE DECISION RULES:
1. If NO evidence FOR anomalous pricing but ANY evidence AGAINST → "No"
2. If evidence FOR anomalous pricing but NO evidence AGAINST → evaluate strength
3. If both types exist → weigh the evidence (prioritize threshold-meeting items)
4. The HEAVILY "for" or "against" far outweighs the just "against" anomalous pricing as those regions are 
highly price informative.
DECISION CRITERIA:
- "Yes": Evidence target is overpriced
- "No": Evidence shows pricing is justified
- "Unsure": Insufficient or conflicting evidence
Format response as JSON: {"explanation": "reasoning (80 words max)", "decision": "Yes/No/Unsure"}"""
\end{verbatim}
\end{framed}

\begin{figure}
    \centering
    \includegraphics[height=\linewidth, angle=-90]{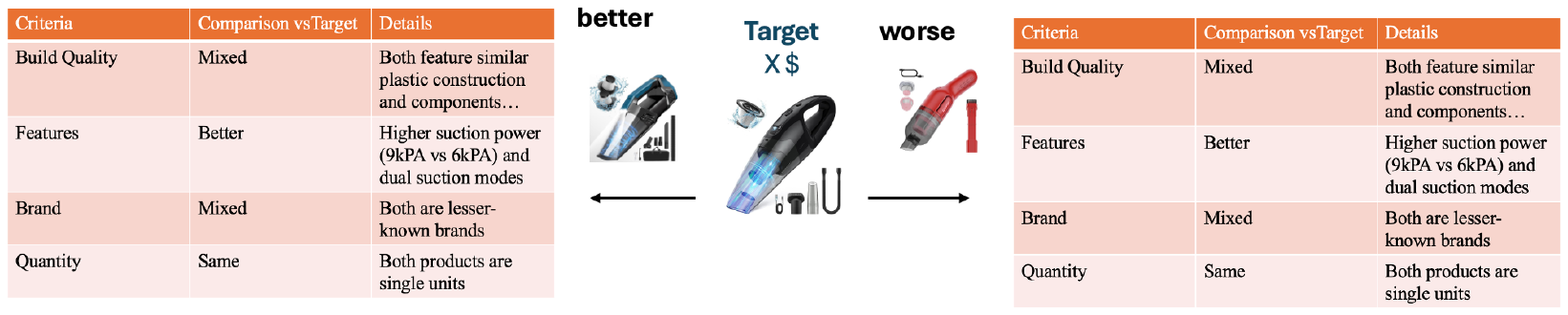}
    \caption{Example for the Utility Agent analysis.}
    \label{fig:sample}
\end{figure}

Figure~\ref{fig:sample} is the sample of Utility analysis for worse, mixed, better buckets. 

As an explanation of the decision agent, consider a \$150 wireless mouse. Under Worse-Pricier Veto, finding a single inferior mouse priced at \$180 would classify our target as not anomalous. Under Quadrant Voting, if we find three better mice at \$100 and two worse mice at \$200, the target would be flagged as anomalous since better-cheaper products outnumber the worse-pricier ones.

\subsection{Different LLM performances}
To compare different LLM models refer to Table~\ref{tab:llm_config}. The same prompt templates were used across all models without modification to ensure fair comparison.
\begin{table*}[ht]
\centering
\small 
\setlength{\tabcolsep}{4pt} %
\begin{tabular}{lcccccccc}
\hline
\textbf{Configuration} & \textbf{SS F1}$\uparrow$ & \textbf{SS Pre.}$\uparrow$ & \textbf{SS Rec.}$\uparrow$& \textbf{Agreement}$\uparrow$ &\textbf{EC F1}$\uparrow$ & \textbf{EC Pre.}$\uparrow$ & \textbf{EC Rec.}$\uparrow$& \textbf{Outlier Rate} \\
\hline
Amazon Nova Pro & 0.46 & 0.3 & 1.0 & 3.0\%&0.29& 0.16& 0.28 & 26.2\%\\
Claude 3.5 Sonnet& 0.67 & 0.54 & 0.88 & 42.1\%&0.50& 0.33& 1.0&10.4\% \\
Claude 4 Sonnet& 0.54& 0.39 & 0.88 & 21.6\% &0.33& 0.20& 1.0&24.2\% \\
\hline
\end{tabular}
\caption{Performance comparison across different LLM models. Padding and Neighbor are set to LLM-Based and 7, respectively. }
\label{tab:llm_config}
\end{table*}

\subsection{Computational Cost and Scaling}
\label{ss:compute_and_scaling}
We can calculate how the model is scalable and cost efficient in large scale using the following equation:
\[
\text{Cost/item} = \frac{\text{API tokens} \times \text{rate}}{\text{batch size}}
\]
The estimated cost summary is presented in Table~\ref{tab:cost_comparison}. Here, if we consider human hourly wage as low as 10\$ and they can audit 3 products per hour, Claude can audit about 37 products per hour and the cost is 3.90\$ per hour. Moreover, there are also potential optimizations like caching, parallel batching, and neighbor reuse to lower the LLM cost.
\begin{table}[htbp]
\centering

\begin{tabular}{ccccc}
\hline
\textbf{\# of Products} & \textbf{Time (Claude 3.5)} & \textbf{Time (Human)} & \textbf{Cost (Claude)} & \textbf{Cost (Human)} \\
\hline
10 & 0.27 & 3.33 & \$1.05 & \$33.33 \\
\hline
1,000 & 27.03 & 333.33 & \$105.30 & \$3,333.33 \\
\hline
400,000,000 & 10,810,810.81 & 133,333,333.33 & \$42,120,000.00 & \$1,333,333,333.33 \\
\hline
\end{tabular}
\caption{Comparison of Analysis Costs: Claude 3.5 vs. Human. Time is measured in hours.}
\label{tab:cost_comparison}
\end{table}

\end{document}